\documentclass[times, review, 10pt]{elsarticle}

\usepackage{graphics} 
\usepackage{epsfig} 
\usepackage{multirow} 
\usepackage{caption}
\usepackage{xcolor}
\usepackage{adjustbox}
\usepackage{booktabs}
\usepackage{wrapfig}
\usepackage[ruled,linesnumbered]{algorithm2e}
\usepackage{amsmath} 
\usepackage{amssymb}  
\usepackage{hyperref}
 
\usepackage{cleveref}
\Crefname{figure}{Fig}{Figures}

\journal{}

\begin{document}

\begin{frontmatter}



\title{SceneLLM: Implicit Language Reasoning in LLM for Dynamic Scene Graph Generation}

\author[label1]{Hang Zhang}
\affiliation[label1]{organization={Information Systems Technology and Design Pillar},
            addressline={Singapore University of Technology and Design}, 
            city={Singapore},
            postcode={487372}, 
            country={Singapore}}

\affiliation[label2]{organization={School of Computing and Communications},
            addressline={Lancaster University},
            city={Lancaster},
            postcode={LA1 4YW},
            country={United Kingdom}}

\author{Zhuoling Li\corref{cor1}\fnref{label2}}
\author[label2]{Jun Liu}
\cortext[cor1]{Corresponding Author: \href{mailto:z.li81@lancaster.ac.uk}{z.li81@lancaster.ac.uk}
}


\begin{abstract}
Dynamic scenes contain intricate spatio-temporal information, crucial for mobile robots, UAVs, and autonomous driving systems to make informed decisions. Parsing these scenes into semantic triplets $<$Subject-Predicate-Object$>$ for accurate Scene Graph Generation (SGG) is highly challenging due to the fluctuating spatio-temporal complexity. Inspired by the reasoning capabilities of Large Language Models (LLMs), we propose \textit{SceneLLM}, a novel framework that leverages LLMs as powerful scene analyzers for dynamic SGG. Our framework introduces a Video-to-Language (V2L) mapping module that transforms video frames into linguistic signals (scene tokens), making the input more comprehensible for LLMs. To better encode spatial information, we devise a Spatial Information Aggregation (SIA) scheme, inspired by the structure of Chinese characters, which encodes spatial data into tokens. Using Optimal Transport (OT), we generate an implicit language signal from the frame-level token sequence that captures the video’s spatio-temporal information. To further improve the LLM’s ability to process this implicit linguistic input, we apply Low-Rank Adaptation (LoRA) to fine-tune the model. Finally, we use a transformer-based SGG predictor to decode the LLM's reasoning and predict semantic triplets. Our method achieves state-of-the-art results on the Action Genome (AG) benchmark, and extensive experiments show the effectiveness of \textit{SceneLLM} in understanding and generating accurate dynamic scene graphs.
\end{abstract}




\end{frontmatter}

\section{Introduction}
\label{sec1}

\begin{figure}[h]
    \centering
    \includegraphics[width=\linewidth]{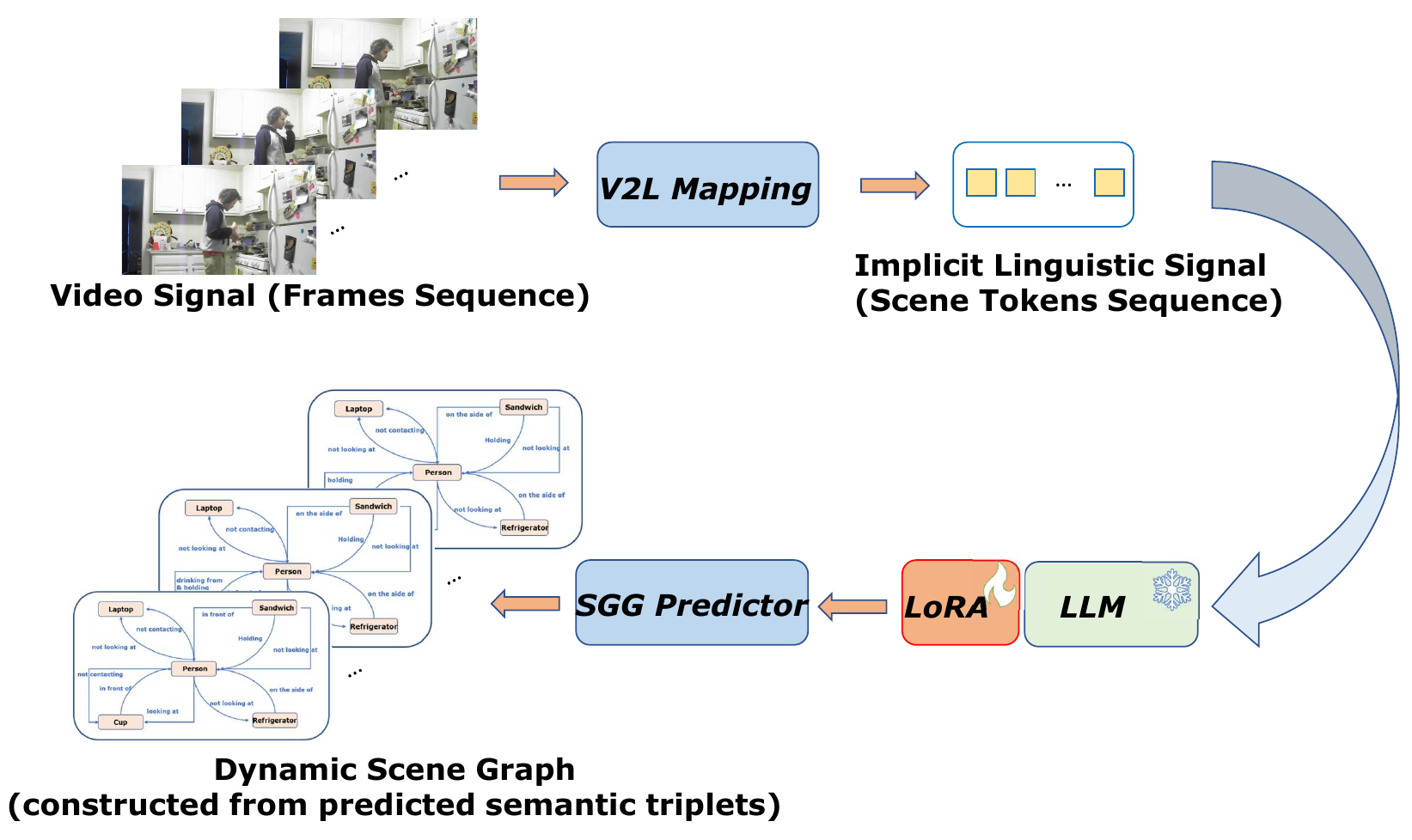}
    \caption{Overview of our proposed SceneLLM framework. In our framework, given an input video signal, we first conduct a Video-to-Language (V2L) Mapping process to obtain the corresponding implicit linguistic signal which is more friendly to LLM. We then perform implicit language reasoning via the fine-tuned LLM with LoRA and employ the SGG predictor to generate the dynamic scene graph.}
    \label{fig:overview}
\end{figure}

A visual scene can be parsed into a semantic structured scene graph in which the semantic entities are the nodes and the relationships containing spatial-temporal information are the edges linking the nodes \cite{xu2017scenegraph}. Dynamic Scene Graph Generation (SGG) involves the spatial localization of objects and inference of semantic predicates among objects so that it can deliver fine-grained semantic information efficiently which is highly beneficial for complex dynamic scene understanding \cite{tr2, Pu2024STKET}. Hence, scene graphs have been commonly applied to aid safe and reasoned decision-making and behavior planning for kinds of robots and autonomous systems \cite{amiri2022reasoning,jiao2022sequential}. 

Compared to static image SGG, the intricate spatio-temporal motion interaction among objects and model inference bias caused by the long-tail distribution of the dataset, place the task of dynamic SGG in a more challenging position.
Even though the promising application prospects of dynamic SGG have spurred a multitude of related exertions and constructive works in terms of spatial-temporal modeling \cite{TRACE, tr2, Pu2024STKET, lin2020gps, Cong_2021_ICCV,li2022dynamic} and unbiased design\cite{ Lin_Shi_Zhan_Yang_Wu_Tao_2024, nag2023unbiased,wang2023triple}, significantly advancing the field of scene understanding, dynamic SGG is still an arduous yet immaturely addressed task \cite{Pu2024STKET}.

Over the years, the advent of the Large Language Models (LLMs) has injected new vitality into all walks of life involving AI. LLMs such as GPT\cite{achiam2023gpt} and LLaMA\cite{touvron2023llama} have achieved tremendous success not only in the field of Natural Language Processing (NLP) but also in various non-text tasks \cite{wang2024survey,wang2024visionllm}. Benefiting from training on massive text corpus which contains tons of visual scene descriptions, LLMs naturally possess rich implicit knowledge on visual world\cite{minaee2024large, hadi2023survey}. Moreover, several recent compelling studies have demonstrated that LLMs can effectively model and comprehend the visual world from textual descriptions and perform dependable reasoning \cite{sharma2024vision,ghanimifard2019neural,huang-chang-2023-towards, feng2024layoutgpt, zhang2024enhancing}. LayoutGPT \cite{feng2024layoutgpt} operates language models to make spatial arrangements, thereby showing the powers of LLMs in visual planning. Ghanimifard et al.\cite{ghanimifard2019neural} indicate the language model’s ability to distinguish functional and geometric biases of spatial relations, despite lacking visual scene features. Zhang et al.\cite{zhang2024enhancing} demonstrate the ability of LLMs to judge the rationality of dynamic spatio-temporal relationships via structured text prompts. Inspired by the above research, we are pushed to wonder such an interesting question as the following: 

\textit{Can LLM be regarded as a scene analyzer for reasoning semantic relationships within scenes, given its inherent rich implicit knowledge and powerful capability of modeling the visual world?} 

To answer the above question, we propose a novel dynamic SGG framework named \textbf{\textit{SceneLLM}} as shown in Fig.\ref{fig:overview}. Our SceneLLM is mainly comprised of three parts: \textit{i)} the Video-to-Language (V2L) Mapping part, \textit{ii)} LLM fine-tuning and implicit reasoning part, and \textit{iii)} SGG preditor. First of all, the LLM was trained on a text corpus and fine-tuned with language instructions, both of which exhibit discreteness and hierarchy.
Hence, the video signal, which is significantly different from the language signal, is extremely unfriendly and incompatible for LLMs to understand. Therefore, we devised a V2L Mapping process to transform the video signal into a language-like signal. During the transformation process, inspired by the representation of Chinese characters \cite{wang2021improving}, to encode necessary spatial-temporal information into implicit language signals, the Spatial Information Aggregation (SIA) and Optimal Transport (OT) strategies have been included. Notably, transformation is carried out in the form of embedding features, hence the output of the transformation is implicit linguistic tokens rather than explicit text. Furthermore, we fine-tune the LLM with pre-trained weights via LoRA \cite{hu2021lora} to help LLM better understand the implicit language-like instruction and conduct LLM's reasoning. It is important to note that we do not directly obtain explicit semantic triplets from LLM, as the input is an implicit embedding resembling language. Instead, we obtain embeddings containing spatial-temporal semantics after LLM inference, which we refer to as implicit inference. Hence, one simple transformer-based SGG predictor is employed to decode the reasoned embeddings of LLM and obtain semantic triplets to form dynamic scene graphs.

In summary, our contributions are as follows: (1) We propose a novel dynamic SGG framework named SceneLLM. To the best of our knowledge, this work is the first to consider LLM as a scene analyzer through implicit language reasoning. (2) We devise a V2L Mapping process to transform the video signal into a language-like signal as the friendly and compatible input of LLM. (3) SceneLLM achieves state-of-the-art performance on the mainstream Action Genome (AG) \cite{ji2020action} evaluation benchmark.

\section{RELATED WORK}

\subsection{Scene Graph Generation (SGG)}
The scene graph represents a typical semantic graph structure that captures the relationships between objects within a scene, facilitating understanding and representation of visual information in a structured form \cite{xu2017scenegraph}. Since its initial proposal for image retrieval tasks in \cite{johnson2015image}, scene graphs have garnered significant interest in computer vision, leading to numerous advancements aimed at enhancing performance in tasks like visual question answering (VQA), image and video captioning, retrieval, and generation \cite{qian2022scene,zhai2024commonscenes,cong2023ssgvs, Divide-and-Conquer, TANG2022108792, M3Net}. Recently, scene graph generation (SGG) has become particularly impactful in robot planning tasks, where visual scene comprehension is crucial. For instance, \cite{jiao2022sequential} highlights SGG's role in robotic action sequencing, while \cite{amiri2022reasoning} addresses robot planning under partial observability by integrating local scene graphs for improved global task execution. Additionally, \cite{zeng2018semantic} introduces a goal-oriented Programming by Demonstration approach, employing scene graphs to treat the robot as a functional operator. Consequently, the quest for high-quality image and video scene graph generation has become a focal area of research.

In \textbf{Image SGG}, various methods have been developed to enhance scene graph precision. For instance, \cite{zhang2019graphical} utilizes contrastive loss to refine predicate recognition, while \cite{xu2017scenegraph} and \cite{zellers2018neural} incorporate spatial context to improve model performance. Advanced architectures like the encoder-decoder model by \cite{cong2023reltr} and the two-stage transformer approach by \cite{kundu2023ggt} further push the boundaries of relational reasoning. \textbf{Dynamic SGG} has also progressed, with methods such as spatial-temporal modeling \cite{Cong_2021_ICCV} and cross-modal knowledge distillation \cite{tr2} enhancing time-dependent relation identification. Addressing model bias has also become significant, with techniques like prior knowledge integration \cite{Pu2024STKET} and memory-guided training \cite{nag2023unbiased}. Our approach diverges by leveraging large language models (LLMs) to infer semantic relationships and generate highly detailed, dynamic scene graphs.

\subsection{Large Language Models (LLMs)}
Recently, increasingly powerful large language models (LLMs) like GPT-4 \cite{achiam2023gpt} and LLaMA3 \cite{touvron2023llama} have emerged, achieving remarkable success across various NLP tasks. These tasks include question answering \cite{zhu2021retrieving}, multilingual translation \cite{he2024exploring}, and so on. Benefiting from pre-training on vast text corpora, LLMs have developed a rich repository of tacit knowledge \cite{ouyang2022training}, enabling them to perform well even outside traditional NLP tasks. In computer vision, for example, LLMs are making strides in scene understanding and visual representation \cite{wang2024visionllm,gong2024llms}. Furthermore, a detailed survey \cite{huang-chang-2023-towards} highlights that LLMs display enhanced reasoning abilities as model size increases. A recent study \cite{kevian2024capabilities} also explores LLMs’ potential in control engineering, a domain that demands complex integration of mathematical and engineering principles. Building on these developments, our work leverages the robust reasoning capabilities of LLMs to advance scene understanding tasks, tapping into their broad knowledge and adaptability.

\section{METHOD}
\begin{figure}[ht]
    \centering
    \includegraphics[width=\textwidth]{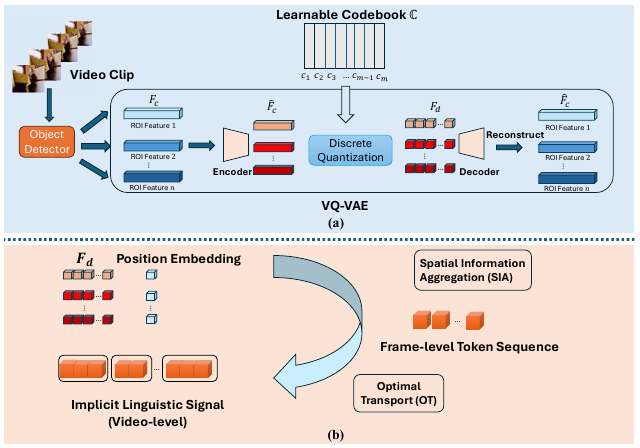}
    \caption{Video-to-Language (V2L) Mapping Process: (a) Features of the regions of interest (ROIs) in video frames are extracted via objector detector, and then the learned VQ-VAE discretely quantizes the extracted ROIs features. (b) SIA and OT schemes to embed spatial-temporal information into implicit linguist tokens which are used as inputs of LLM.}
    \label{fig:V2L}
\end{figure}

\textbf{Problem Statement.} Dynamic SGG aims to generate a set of structured scene representations \(G = \{G_\tau\}_{\tau=1}^{T}\) for a video $V$ with total $T$ frames. Each scene graph $G_\tau$ corresponding to the frame $I_{\tau}$ consists of multiple triplets $\{S_\tau, P_\tau, O_\tau\}$ ($<$Subject-Predicate-Object$>$), in which $S_\tau=\{S_1^{\tau}, S_2^{\tau},$ $\cdots, S_{n(\tau)}^{\tau}\}$ and $O_\tau=\{O_1^{\tau}, O_2^{\tau},\cdots, O_{n(\tau)}^{\tau}\}$ are the detected objects in the frame $I_\tau$, and $P_\tau=\{P_1^{\tau}, P_2^{\tau},\cdots, P_{m(\tau)}^{\tau}\}$ contains relationships among the detected objects, also referring to nodes $N_\tau$ and edges $E_\tau$ in the scene graph $G_\tau$ respectively. Besides, all detected objects have corresponding bounding boxes $B_\tau=\{B_1^{\tau}, B_2^{\tau},\cdots, B_{n(\tau)}^{\tau}\}$. $S_\tau$ and $O_\tau$ are the subsets of the set of object categories $Y_o$, and $P_\tau$ is also the subset of the set of predicate categories $Y_p$. 

\subsection{The Overall Framework} 

The overall structure of our proposed SceneLLM framework is shown in Fig.\ref{fig:overview}. First of all, the V2L mapping module in SceneLLM will map the input video signal (the frame sequence) into an implicit linguistic signal (dubbed "scene sentences") which focuses on transferring spatial-temporal information via SIA and OT schemes. Afterward, SceneLLM feeds the "scene sentences" into LLM to implicitly reason about the corresponding semantic relationships existing in the scene. Finally, we employ a straightforward SGG predictor to decode the reasoned output and produce semantic triplets. We will illustrate each of the key components in SceneLLM in the following subsections.

\subsection{Video-to-Language (V2L) Mapping Module}

As shown in Fig.\ref{fig:V2L}, V2L Mapping has two stages. (a) We first learn an object-oriented vector quantized variational autoencoder (VQ-VAE \cite{van2017neural}) to discretize features of objects in the scene. (b) Then aggregate the discrete feature tokens with spatial information with the SIA scheme to obtain frame-level token sequence, thereby an optimal transport scheme to construct a language-like hierarchy with temporal information while forming LLM-friendly "scene sentences". 

\textbf{Feature Discrete Quantization.} Specifically, at the time $\tau$, the object detector detects objects in frame $I_\tau$ and outputs the corresponding bounding boxes $B_\tau$, categories, and visual features $F_c$ of the objects (ROI Features). Next, our VQ-VAE consisting of an encoder $\mathbb{E}$, a decoder $\mathbb{D}$, and a codebook $\mathbb{C}=\{c_k\}_{k=1}^m$ ($c_k\in\mathbb{R}^l$, where dimension $l$ is the same as the dimension of the word tokens in the LLM used) discretizes the visual features of the objects. Given the extracted $n$ visual features $F_c=\{f_{1:n}\}$, the conventional encoder $E$ encodes input features into latent features $\tilde{F}_c=\{\tilde{f}_{1:n}\}$ ($\tilde{f}_n\in\mathbb{R}^l$). To obtain discrete feature tokens $F_d=\{f_{1:n}^d\}$, we conduct a standard discrete quantization operation in VQ-VAE by replacing each latent feature $\tilde{f}_{n}$ with its nearest unit $c_k$ of the codebook $\mathbb{C}$ as follows:
\begin{equation}
\label{eq:quantization}
    f_n^d=\underset{c_k \in \mathbb{C}}{\arg\min}(\|\tilde{f}_n - c_k\|_2)
\end{equation}
Once discrete quantization is completed, the decoder $\mathbb{D}$ will reconstruct the visual features $F_c$ using discrete features $F_d$.

\textbf{Implicit Linguistic Signal Generation.} Once we obtain the discrete features of objects $F_d$, the next step is to generate a linguistic tokens sequence. Considering the spatial information between different objects in the static frame and temporal information in the dynamic scenes (video), the generated linguistic signal should also contain corresponding spatial-temporal information. To this end, inspired by the study \cite{wang2021improving} on the spatial semantic representation of Chinese characters, we propose to aggregate spatial information in the discrete feature $F_d$ with an SIA scheme to obtain a frame-level token and thereby generate a video-level linguistic signal from the frame-level token sequence in an Optimal Transport (OT) \cite{volt,chen2018improving} manner.

\begin{figure*}
    \centering
     \includegraphics[width=\textwidth]{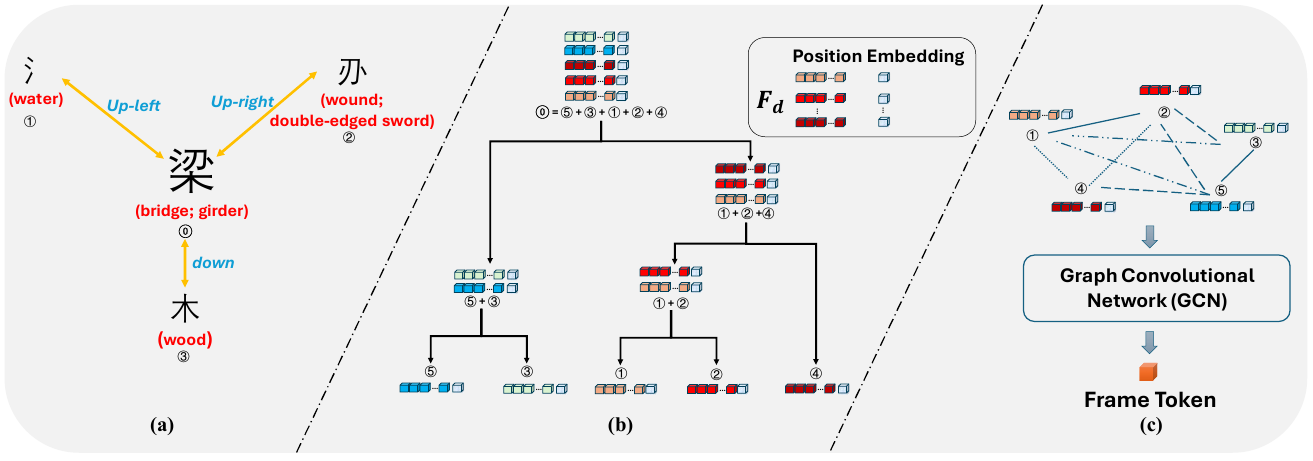}
    \caption{Spatial Information Aggregation for Frame-level Token Generation: (a) Illustration of Spatial Representation of Chinese Characters. (b) Hierarchical Clustering of Discrete Objects' Features for Spatial Correlation. (c) Chinese Character-like Frame-level Token Generation with GCN.}
    \label{fig:sia}
\end{figure*}

\textbf{Spatial Information Aggregation (SIA).} Specifically, as shown in Fig.\ref{fig:sia} (a), the Chinese character is composed of multiple radicals combined in a certain spatial structure (i.e. up and down, left and right, or their combined positions). Each radical stands for a certain object which has a specific meaning. For easy understanding, \textcircled{\raisebox{-0.9pt}{0}} represents the Chinese Character that means bridge or girder. It can be split into 3 radicals, namely \textcircled{\raisebox{-0.9pt}{1}} (water) located at the up-left, \textcircled{\raisebox{-0.9pt}{2}} (sword or wound) positioned at the up-right, and \textcircled{\raisebox{-0.9pt}{3}} (wood) situated at the bottom. In ancient times, people built a bridge \textcircled{\raisebox{-0.9pt}{0}} over rivers \textcircled{\raisebox{-0.9pt}{1}} and cut down wood \textcircled{\raisebox{-0.9pt}{3}} with swords \textcircled{\raisebox{-0.9pt}{2}} to make it. Inspired by this, we treat a single video frame (scene) as a Chinese character, the objects in the video frame or scene as radicals, and the spatial relation of objects as the spatial structure of the Chinese character. What we need to do is to determine the discrete features of objects that belong to which radicals and aggregate them to form a frame-level token, thereby, for a video (frame sequence), generating a frame-level token sequence. To this end, we first embed position information into discrete objects feature of each frame $F_d=\{f_{1:n}^d\}$ by a simple Multilayer Perceptron (MLP) as follows:
\begin{equation}
\label{eq:position_embedding}
\begin{aligned}
    P_n &= MLP([x_n,y_n,w,h])\\
    f_n^{d^+} &= f_n^d \oplus P_n
\end{aligned}
\end{equation}
where $[x_n,y_n,w,h]$ is the position information of the $n$-th object that can be obtained from the object detector, $(x_n,y_n)$ is the center coordinate of the corresponding bounding box, $(w,h)$ are the width and height of the frame respectively. And the $\oplus$ is the vector concatenation operator. 

Then, as Fig.\ref{fig:sia} (b) and (c) show, once we obtain the position-embedded discrete object features $F_d^+=\{f_{1:n}^{d^+}\}$, to figure out the spatial relation between objects like positions of different radicals, a Hierarchical Clustering (HC) algorithm is applied. In the context of the scene graph, the objects are nodes $N$ and the relationships between objects are edges $E$. So, the clustering results can indicate the connectivity $C$ of nodes. Next, to produce the frame-level token $t$ ("Chinese Character") for each frame, we leverage a Graph Convolutional Network (GCN) that can effectively represent graph structure to fuse features $F_d^+$ as follows:
\begin{equation}
\label{eq:frame-level-token}
    \begin{aligned}
    C &= HC(F_d^+)  \\
    t &= GCN(F_d^+, C)
    \end{aligned}
\end{equation}
Hence, for a video $V$ with $T$ frames, we will get a frame-level token sequence $\{t_{\tau}\}_{\tau=1}^{T}$.

\begin{figure}[ht]
    \centering
    \includegraphics[width=\linewidth]{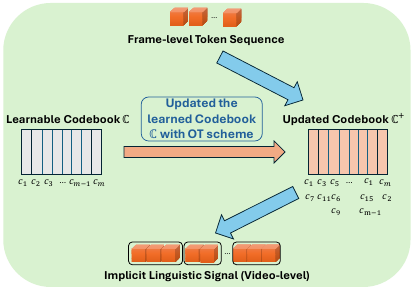}
    \caption{Update for an optimal codebook $\mathbb{C^+}$ via an optimal transport scheme so that dynamic information can be embedded into implicit linguistic signal.}
    \label{fig:OT}
\end{figure}

\textbf{Optimal Transport (OT) Scheme.} 
Since the frame-level token is independent and only contains spatial information of its corresponding frame. To understand complex triplets related to high temporal information in dynamic scenes, we have to further process the frame-level token sequence $\{t_{\tau}\}_{\tau=1}^{T}$ for temporal consistency. For example, the triplet \texttt{<person-hold-cup>} exists in a static frame (scene), while with temporal information (dynamic scenes), it could be \texttt{<person-drinking from-cup>}. To this end, we propose to apply an OT scheme \cite{volt,chen2018improving} to transport semantic information from different frame-level tokens to equip the generated implicit linguistic signal with temporal information as shown in Fig.\ref{fig:OT}. Various units in the learned codebook $\mathbb{C}$ contain different semantic information, our first step is to find out which units should be gathered to make the units in the updated codebook $\mathbb{C^+}$ with the temporal semantic property. From the information theory perspective, gathering discrete units can be viewed as a process of entropy reduction because of the lower uncertain information in codebook $\mathbb{C^+}$. On the other hand, considering the semantic representation capability of the codebook $\mathbb{C^+}$, a reasonable codebook size is also important. In short, we shall update for an optimal codebook $\mathbb{C^+}$ with lower entropy and reasonable size.

Specifically, following the previous works \cite{gong2024llms, nag2023entropy}, we set a fixed size increment $\Delta s$ and define the $k$-th codebook $\mathbb{C}^+_k$ as the codebook with $k\times \Delta s$ units. Then, we try to find out the optimal set of units, where each new unit consists of the previously learned units so that we can formulate this as an OT problem. The entropy of the $k$-th codebook $\mathbb{C}^+_k$ is:
\begin{equation}
    \label{eq:entropy}
    \begin{aligned}
    H_{\mathbb{C}_k^+}&=  -\sum_{u_j^+ \in \mathbb{C}_k^+} P\left(u_j^+\right) \log P\left(u_j^+\right) \\
    & =-\sum_{u_j^+ \in \mathbb{C}_k^+} \sum_{u_i \in \mathbb{C}} P\left(u_j^+, u_i\right) \log P\left(u_j^+\right) \\
    & =-\sum_{u_j^+ \in \mathbb{C}_k^+} \sum_{u_i \in \mathbb{C}} P\left(u_j^+, u_i\right) \log P\left(u_j^+, u_i\right) \\
    & \quad -\sum_{u_j^+ \in \mathbb{C}_k^+} \sum_{u_i \in \mathbb{C}} P\left(u_j^+, u_i\right)\left( -\log \frac{P(u_j^+, u_i)}{P(u_j^+)} \right)\\
    & =-\sum_{u_j^+ \in \mathbb{C}_k^+} \sum_{u_i \in \mathbb{C}} P\left(u_j^+, u_i\right) \log P\left(u_j^+, u_i\right) \\
    & \quad -\sum_{u_j^+ \in \mathbb{C}_k^+} \sum_{u_i \in \mathbb{C}} P\left(u_j^+, u_i\right)\left(-\log P\left(u_i \mid u_j^+\right)\right)
    \end{aligned}
\end{equation}

where $P(u_j^+)$ is the relative frequency of the $j$-th unit $u_j^+$ from the codebook $\mathbb{C}^+$ and $P(u_i \mid u_j^+)$ is the probability of the learned unit $u_i$ appearing in the new unit $u_j^+$.

Next, to formulate the objective function that minimizing the entropy of $\mathbb{C}_k^+$, following \cite{gong2024llms,peyre2019computational,volt}, we first set the transport matrix $P\in\mathbb{R}^{m \times (k\cdot\Delta s)}$ representing the allocation of the learned units $u$ to the new unit $u^+$, where the $(j, i)$-th element is $P(u_j^+, u_i)$, the distance matrix $D\in\mathbb{R}^{m \times (k\cdot\Delta s)}$ representing the cost of transport, where the $(j, i)$-th element is $\log P\left(u_i \mid u_j^+\right)$, and the $H(P)$ representing the entropy of the probability distribution $P(u_j^+, u_i)$. Therefore, according to Eq.\ref{eq:entropy}, our objective function is:

\begin{equation}
\label{eq:obj_func}
\begin{aligned}
\underset{\boldsymbol{P} \in \mathbb{R}^{m \times(k \cdot \Delta s)}}{\arg \min } H(\boldsymbol{P})+\sum_j \sum_i \boldsymbol{P}(j, i) \boldsymbol{D}(j, i) 
\end{aligned}
\end{equation}

Then, for this OT problem that finds the optional way to transfer units in codebook $\mathbb{C}$ to form an updated codebook $\mathbb{C^+}$, we utilize the Sinkhorn algorithm \cite{peyre2019computational,volt} to optimize the defined object function efficiently. The overall algorithm for obtaining the updated codebook $\mathbb{C}^+$ is shown in Algorithm \ref{algo:update-codebook}. Finally, once obtaining the optimal candidate codebook $\mathbb{C^+}$, similar to \cite{gong2024llms}, we generate the implicit linguistic signal $\mathcal{S}_{LLM}$ via an auto-regressive model $\mathcal{M}$ with the updated codebook $\mathbb{C^+}$. Please refer to Section 4.2 for more details about $\mathcal{M}$.

\begin{algorithm}[ht]
\caption{Update Codebook}
\label{algo:update-codebook}
\SetKwInOut{Input}{Input}\SetKwInOut{Output}{Output}
\Input{learned codebook $\mathbb{C}$, initial step $k=0$, increment $\Delta s$, 
        initial updated codebook $\mathbb{C}_0^+=\mathbb{C}$, candidate codebooks $L$ = [$\mathbb{C}_0^+$],  \textit{Flag}$=$\texttt{True}}
\Output{Optimal Updated Codebook $\mathbb{C}^+$}

\While{Flag}{$k\leftarrow k+1$\;
 Expand codebook size to $k\times \Delta s$\;
 Construct $\mathbb{C}_k^+$ based on optimizing Eq.\ref{eq:obj_func} with the OT scheme\;
 Calculate $H_{\mathbb{C}_k^+}$ (entropy of $\mathbb{C}_k^+$) based on Eq.\ref{eq:entropy}\;
 $L \leftarrow L+[\mathbb{C}_k^+]$\;
 Calculate entropy decrease $\Delta H_k \leftarrow H_{\mathbb{C}_k^+}-H_{\mathbb{C}_{k-1}^+}$\;
\If{$\Delta H_k > \Delta H_{k-1}$}{\textit{Flag}$=$\texttt{Flase}\;}
$\mathbb{C}^+\leftarrow \mathbb{C}^+_{k-1}$ (from $L$)\;}
\end{algorithm}

\subsection{Implicit Language Reasoning in LLM}

So far, we have acquired the implicit linguistic signal $\mathcal{S}_{LLM}$ which encompasses the spatio-temporal semantic information of the scene. This signal can be regarded as a "scene sentence" or a "scene token sequence". To parse the semantic relationships within the scene, we leverage the implicit reasoning capabilities of LLM as it has been proven its rich implicit knowledge can be used to reason and model the visual world \cite{zhang2024enhancing,feng2024layoutgpt,sharma2024vision,ghanimifard2019neural,huang-chang-2023-towards}. Specifically, we designed the following prompt as LLM's input:

\begin{quote}
    \texttt{Given such a scene sentence [$\mathcal{S}_{LLM}$], please parse the \\
    relationships between the person and objects in the scene.}

\end{quote}

Besides, to help the LLM better understand our "scene sequence", we also execute LoRA \cite{hu2021lora} to fine-tune LLM. Note that we do not need the explicit output of LLM, instead, the final scene graph will be generated by a simple transformer-based SGG predictor $\mathcal{D}_{SGG}$ \cite{Cong_2021_ICCV}, and the input of $\mathcal{D}_{SGG}$ is the reasoned feature output of LLM. The whole decoding process is as follows:
\begin{equation}
\label{eq:decoding}
\begin{aligned}
    F_{implicit}&=LLM(\mathcal{S}_{LLM})\\
    \hat{Y}&=\mathcal{D}_{SGG}(F_{implicit})
    \end{aligned}
\end{equation}
where the $F_{implicit}$ is the hidden feature of LLM's final block and $\hat{Y}$ is the generated scene graph consisting of multiple triplets.

\subsection{Training and Inference}
\textbf{Inference.} 
Given a video $V$ with T frames, we first extract the visual features $F_C$ of objects in each frame via a pre-trained object detector. Then, $F_C$ is encoded to latent features $\tilde{F_C}$ via the encoder $\mathbb{E}$ of our VQ-VAE, and $\tilde{F_C}$ is quantized to $F_D$ by the learned codebook $\mathbb{C}$. After that, we encode spatial information into $F_D$ via MLP to obtain $F_D^+$ (Eq. \eqref{eq:position_embedding}), and aggregate $F_D^+$ to form the frame-level token sequence $t$ via GCN (Eq. \eqref{eq:frame-level-token}). Subsequently, we transform the frame-level token sequence $\{t_{\tau}\}_{\tau=1}^{T}$ to implicit linguistic signal $\mathcal{S}_{LLM}$ via the OT scheme (Eq. \eqref{eq:obj_func}). Finally, we feed $\mathcal{S}_{LLM}$ into the LLM to reason about the semantic relationships in the scene and generate the scene graph via the SGG predictor $\mathcal{D}_{SGG}$.

\textbf{Training.} 
Our framework is optimized in two stages: the VQ-VAE training and the SceneLLM training. At the first stage, we pre-train the VQ-VAE model to encode and quantize visual features of objects. To achieve this, similar to \cite{van2017neural}, VQ-VAE is optimized to reconstruct visual features via encoding, quantization, and decoding process. The VQ-VAE loss function consists of 3 sub-losses, namely reconstruction loss for encoder-decoder learning, embedding loss for codebook learning, and commitment loss for limiting arbitrary growth of codebook volume. The loss function is defined as follows:
\begin{equation}
\label{eq:loss_vqvae}
\begin{aligned}
    Loss_{vae} &= \| F_c - \hat{F}_c \|_2^2 && \text{(recon. loss)}\\ 
    &+ \| sg[\tilde{F}_c] - c \|_2^2 && \text{(embed. loss)}\\
    &+ \lambda \| \tilde{F}_c - sg[c] \|_2^2 && \text{(commit. loss)}
\end{aligned}
\end{equation}
where $sg[\cdot]$ is the stop-gradient operator (i.e. identity function) with zero partial derivatives. 

After VQ-VAE training, we train our SceneLLM to generate scene graphs. 
To reduce the difficulty of fine-tuning the LLM, we adopt a progressive unfreezing training approach. Specifically, freezing VQ-VAE's encoder and LLM, we first optimize the MLP, GCN, and SGG predictor for 30,000 iterations. Then, we equip the LLM with LoRA and fine-tune the MLP, GCN, LLM, and SGG predictor for 50,000 iterations. The loss function for training SceneLLM is defined as follows:
\begin{equation}
    \label{eq:loss_sgg}
    \begin{aligned}
        Loss_{sgg} &= \alpha L_{obj}  + L_{rel}
    \end{aligned}
    \end{equation}
where $L_{obj}$ denotes cross-entropy loss to measure the prediction of entity classification, $L_{rel}$ is binary cross-entropy loss which measures the prediction of relationship classification, and $\alpha$ is the weighting factor.


\section{EXPERIMENTS}
\subsection{Experiment Setting}
\textbf{Dataset.}The \textbf{Action Genome (AG)} dataset \cite{ji2020action} is an extensive video dataset meticulously annotated with frame-level scene graphs. Designed for detailed scene understanding, the dataset contains 36 object classes, with approximately 1.7 million annotated object instances distributed across 26 predicate classes. These annotations provide a thorough exploration of 157 distinct triplet categories, mapped across nearly 234,000 frames, which significantly enhances the dataset’s utility for comprehensive scene analysis. Predicates within the AG dataset are thoughtfully divided into three types: \textbf{(1) Attention Predicates}, which indicate a subject’s focus on objects, \textbf{(2) Spatial Predicates}, which denote relative positioning, and \textbf{(3) Contact Predicates}, capturing the nature of interaction between objects, covering both contact and non-contact forms. This structured approach to predicate categorization enables a nuanced and systematic understanding of the complex relationships within scenes, making the AG dataset highly valuable for tasks involving relational dynamics in videos.

\textbf{Evaluation Tasks and Metrics.} Following \cite{ji2020action, Cong_2021_ICCV}, SceneLLM undergoes evaluation in three tasks: \textbf{(1) Predicate Classification (PREDCLS)}: providing objects labels and bounding boxes of objects to predict predicate labels of object pairs. \textbf{(2) Scene Graph Classification (SGCLS)}: classifying the given ground-truth bounding boxes and predicting predicate labels. \textbf{(3) Scene Graph Detection (SGDET)}: detecting bounding boxes and labels of objects and predicting predicate labels. For metrics, we keep in line with \cite{ji2020action, Cong_2021_ICCV} to apply the widely-used \textbf{Recall@K (K=[10,20,50])} metric to quantify the performance of models for all tasks, which indicates the ratio of ground-truth triplets identified within the top K confidence predictions. Besides, 2 different graph generation rules \cite{ji2020action} are also adopted to realize comprehensive and fair comparison with other models, namely \textbf{ (1) With Constraint} only allow at most one predicate per object pair, and \textbf{(2) No Constraint} allows per objects pair to have multiple relationships.

\subsection{Implementation Details}
Following OED\cite{Wang_2024_CVPR}, we train a Transformer-based detector for object detection which is initialized from a COCO-pre-trained model and fine-tuned on the AG dataset. We set the latent feature dimension $l$ of VQ-VAE to 512, the number of codebook size $m$ to 512, and the weighting factor $\lambda$ in \ref{eq:loss_vqvae} to 0.02. Following \cite{gong2024llms}, the auto-regressive model $\mathcal{M}$ is implemented as a one Convolutional Gated Recurrent Layer with a kernel size of (1,1). During the first training stage (VQ-VAE training), we optimize VQ-VAE for 300,000 iterations using the AdamW optimizer with an initial learning rate of 3e-4. In Eq. \eqref{eq:frame-level-token}, we employ a bottom-up hierarchical clustering algorithm, with average linkage to measure the distance between between clusters, to construct the spatial relation $C$ between object features $F_d$. In addition, $GCN$ consists of two graph convolutional layers. We use frozen LLaMA-13B \cite{touvron2023llama} as our LLM. During the second training stage (SceneLLM training), we first optimize the MLP, GCN, and SGG predictor for 30,000 iterations with an initial learning rate of 1e-5. Then, for the LoRA process, we fine-tune the LLM using the AdamW optimizer with an initial learning rate of 1e-5 for 50,000 iterations. The weighting factor $\alpha$ in Eq. \eqref{eq:loss_sgg} is set to 0.5. We conduct our experiments on Nvidia A5000 GPU.

\begin{table*}[t]
\vspace*{5pt}
\centering
\caption{Evaluation results of our SceneLLM and baselines in the With Constraints setting}
\label{table-constraint-with}
\tabcolsep=0.07cm
\resizebox{0.85\linewidth}{!}{
\begin{tabular}{cccccccccc}
\toprule
\multirow{2}{*}{Method} & \multicolumn{3}{c}{PREDCLS} & \multicolumn{3}{c}{SGCLS} & \multicolumn{3}{c}{SGDET} \\ 
\cmidrule(r){2-4} \cmidrule(r){5-7} \cmidrule(r){8-10}
& R@10    & R@20    & R@50    & R@10    & R@20   & R@50   & R@10    & R@20   & R@50   \\ 
\midrule
M-FREQ\cite{zellers2018neural} & 62.4 & 65.1 & 65.1 & 40.8 & 41.9 & 41.9 & 23.7 & 31.4 & 33.3 \\
VCTREE\cite{tang2018learning} & 66.0 & 69.3 & 69.3 & 44.1 & 45.3 & 45.3 & 24.4 & 32.6 & 34.7 \\
GPS-Net\cite{lin2020gps} & 66.8 & 69.9 & 69.9 & 45.3 & 46.5 & 46.5 & 24.7 & 33.1 & 35.1 \\
TRACE\cite{TRACE} & 64.4 & 70.5 & 70.5 & 36.2 & 37.4 & 37.4 & 19.4 & 30.5 & 34.1 \\
STTran\cite{Cong_2021_ICCV} & 68.6 & 71.8 & 71.8 & 46.4 & 47.5 & 47.5 & 25.2 & 34.1 & 37.0 \\
APT\cite{li2022dynamic} & 69.4 & 73.8 & 73.8 & 47.2 & 48.9 & 48.9 & 26.3 & 36.1 & 38.3 \\
TEMPURA\cite{nag2023unbiased} & 68.8 & 71.5 & 71.5 & 47.2 & 48.3 & 48.3 & 28.1 & 33.4 & 34.9 \\
TR$^2$\cite{tr2} & 70.9 & 73.8 & 73.8 & 47.7 & 48.7 & 48.7 & 26.8 & 35.5 & 38.3 \\
TD$^2$-Net\cite{Lin_Shi_Zhan_Yang_Wu_Tao_2024} & 70.1 & - & 73.1 & 51.1 & - & 52.1 & 28.7 & - & 37.1 \\
FloCoDe\cite{khandelwal2024flocode} & 70.1 & 74.2 & 74.2 & 48.4 & 51.2 & 51.2 & 31.5 & 38.4 & 42.4 \\
OED\cite{Wang_2024_CVPR} & 73.0 & 76.1 & 76.1 & - & - & - & 33.5 & 40.9 & 48.9 \\
DDS\cite{iftekhar2025dds} & - & - & - & - & - & - & \textbf{36.2} & 42.0 & 47.3 \\
STABILE\cite{zhuang2025spatial} & 68.8& 71.8 &71.8 & 49.6& 50.7 &50.7 & 29.9 &34.2& 35.0 \\
DIFFVSGG\cite{chen2025diffvsgg} &   71.9& 74.5 &74.5 &   52.5& 53.7 &53.7 & 32.8& 39.9& 45.5 \\
Ours & \textbf{74.1} & \textbf{77.8} & \textbf{77.8} & \textbf{53.7} & \textbf{55.0} & \textbf{55.0} & 34.9 & \textbf{43.3} & \textbf{49.5} \\
\bottomrule
\end{tabular}}
\end{table*}

\begin{table*}[t]
\vspace*{5pt}
\centering
\caption{Evaluation results of our SceneLLM and baselines in the No Constraints setting}
\label{table-constraint-no}
\tabcolsep=0.07cm
\resizebox{0.85\linewidth}{!}{
\begin{tabular}{cccccccccc}
\toprule
\multirow{2}{*}{Method} & \multicolumn{3}{c}{PREDCLS} & \multicolumn{3}{c}{SGCLS} & \multicolumn{3}{c}{SGDET} \\ 
\cmidrule(r){2-4} \cmidrule(r){5-7} \cmidrule(r){8-10}
& R@10    & R@20    & R@50    & R@10    & R@20   & R@50   & R@10    & R@20   & R@50   \\ 
\midrule
M-FREQ\cite{zellers2018neural} & 73.4 & 92.4 & 99.6 & 50.4 & 60.6 & 64.2 & 22.8 & 34.3 & 46.4 \\
VCTREE\cite{tang2018learning} & 75.5 & 92.9 & 99.3 & 52.4 & 62.0 & 65.1 & 23.9 & 35.3 & 46.8 \\
RelDN\cite{zhang2019graphical} & 75.7 & 93.0 & 99.0 & 52.9 & 62.4 & 65.1 & 24.1 & 35.4 & 46.8 \\
GPS-Net\cite{lin2020gps} & 76.0 & 93.6 & 99.5 & 53.6 & 63.3 & 66.0 & 24.4 & 35.7 & 47.3 \\
TRACE\cite{TRACE} & 73.3 & 93.0 & 99.5 & 36.3 & 45.5 & 51.8 & 27.5 & 36.7 & 47.5 \\
STTran\cite{Cong_2021_ICCV} & 77.9 & 94.2 & 99.1 & 54.0 & 63.7 & 66.4 & 24.6 & 36.2 & 48.8 \\
APT\cite{li2022dynamic} & 78.5 & 95.1 & 99.2 & 55.1 & 65.1 & 68.7 & 25.7 & 37.9 & 50.1 \\
TEMPURA\cite{nag2023unbiased} & 80.4 & 94.2 & 99.4 & 56.3 & 64.7 & 67.9 & 29.8 & 38.1 & 46.4 \\
TR$^2$\cite{tr2} & 83.1 & 96.6 & 99.9 & 57.2 & 64.4 & 66.2 & 27.8 & 39.2 & 50.0 \\
TD$^2$-Net\cite{Lin_Shi_Zhan_Yang_Wu_Tao_2024} & 81.7 & - & 99.9 & 57.2 & - & 69.8 & 30.5 & - & 49.3 \\
FloCoDe\cite{khandelwal2024flocode} & 82.8 & \textbf{97.2} & 99.9 & 57.4 & 66.2 & 68.8 & 32.6 & 43.9 & 51.6 \\
OED\cite{Wang_2024_CVPR} & 83.3 & 95.3 & 99.2 & - & - & - & 35.3 & 44.0 & 51.8 \\
DDS\cite{iftekhar2025dds} & - & - & - & - & - & - & 37.3 & 43.3 & 51.5 \\
STABILE\cite{zhuang2025spatial} & 80.4 &94.5& 99.6 & 57.7& 66.0 &71.3 & 33.7& 41.2& 47.7 \\
DIFFVSGG\cite{chen2025diffvsgg} & 83.1 &94.5 &99.1 &   60.5 &\textbf{70.5}& \textbf{74.4} & 35.4 &42.5 &51.0 \\
Ours & \textbf{83.7} & \textbf{97.2} & \textbf{99.9} & \textbf{61.2} & 69.7 & 71.1 & \textbf{37.4} & \textbf{46.1} & \textbf{53.4} \\
\bottomrule
\end{tabular}}
\end{table*}

\subsection{Performance Results and Comparison}
In Table \ref{table-constraint-with} and Table \ref{table-constraint-no}, we present a detailed comparison of the performance of our proposed SceneLLM model alongside several existing methods on the AG dataset in both \textbf{With Constraint} and \textbf{No Constraint} settings. To provide clarity, the best performance values in each category are highlighted in bold. The results clearly demonstrate that SceneLLM, leveraging the capabilities of a large language model (LLM), consistently achieves state-of-the-art results across all tasks and metrics. This indicates the effectiveness of SceneLLM’s architecture in handling the complexities of scene understanding and generating high-quality relationships within the data. 

For instance, under the \textbf{With Constraint} setting, our SceneLLM outperforms the second-best method, OED, by an average of \textbf{1.5\%} in the PREDCLS task. Specifically, SceneLLM achieves improvements of \textbf{1.1\%} in R@10, \textbf{1.7\%} in R@20, and \textbf{1.7\%} in R@50, showcasing its ability to capture and predict subject-predicate-object relationships with higher accuracy. Additionally, SceneLLM also surpasses OED in SGDET, with an average improvement of \textbf{1.5\%} (\textbf{1.4\%} for R@10, \textbf{2.4\%} for R@20, and \textbf{0.6\%} for R@50), indicating superior detection and graph completion capabilities. In the \textbf{No Constraint} setting, SceneLLM similarly demonstrates robust performance, achieving an average improvement of \textbf{1.0\%} in the PREDCLS task (\textbf{0.4\%} for R@10, \textbf{1.9\%} for R@20, and \textbf{0.7\%} for R@50) and an even more substantial improvement of \textbf{1.9\%} in SGDET (\textbf{2.1\%} for R@10, \textbf{2.1\%} for R@20, and \textbf{1.5\%} for R@50). These results reinforce SceneLLM’s generalization capabilities and adaptability in unconstrained scenarios, further solidifying its place as the leading method in scene graph generation and relationship prediction.

All these results demonstrate the efficacy of our SceneLLM, and prove that LLM can effectively reason semantic relationships within scenes by leveraging its inherent rich implicit knowledge.

\subsection{Visualization}
\begin{figure}[ht]
    \centering
    \includegraphics[width=\linewidth]{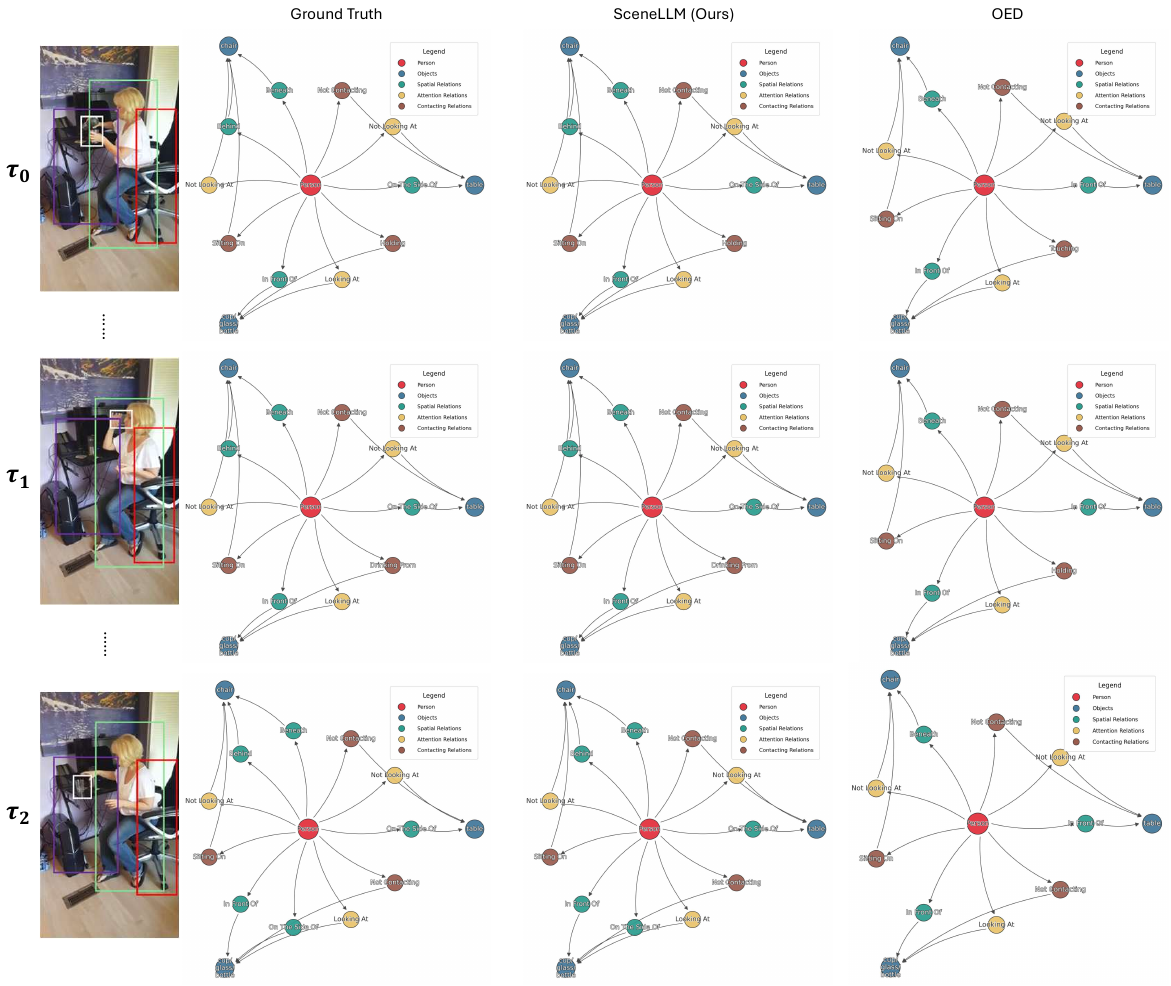}
    \caption{Visualization results of our method. All results are given under the SGDET setup. (Zoom in for the best view.)}
    \label{fig:vis}
\end{figure}
As shown in Fig.\ref{fig:vis}, SceneLLM's visual detection results in dynamic scenes highlight its superior ability to model spatiotemporal relationships. This advancement is the result of SceneLLM's strong inference capabilities on fine-grained spatiotemporal dependencies and the vast and rich pre-trained knowledge contained in LLMs, which enables the ability to accurately parse complex real-world scenes, even as object-to-human interactions dynamically change over time.

\subsection{Ablation Studies}
In this section, we conduct extensive ablation experiments on the AG dataset to investigate our SceneLLM.

\begin{figure}[ht]
    \centering
    \includegraphics[width=0.85\linewidth]{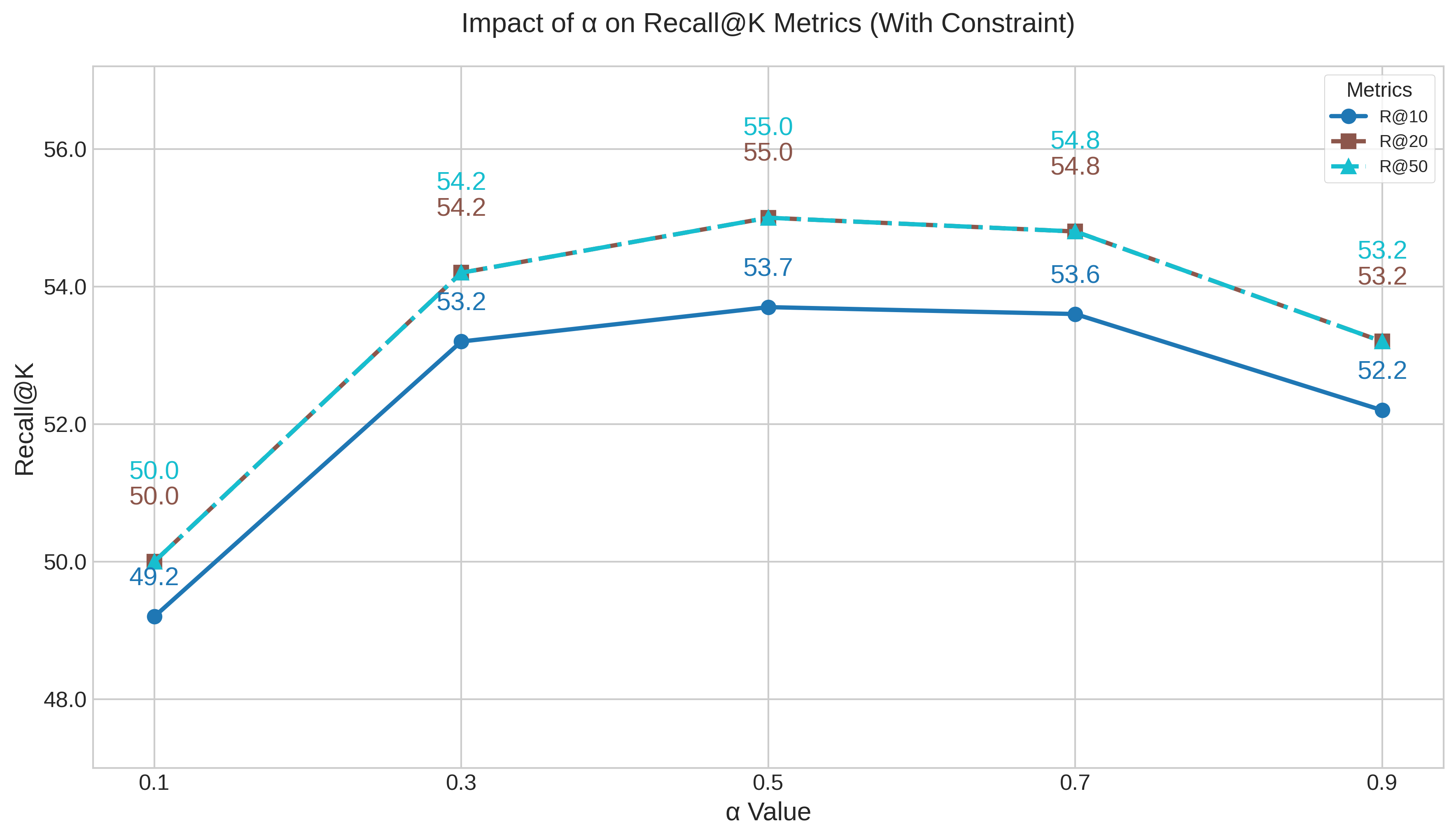}
    \caption{Impact of weight factor $\alpha$ (With Constraint)}
    \label{fig:alpha_w}
\end{figure}

\begin{figure}[ht]
    \centering
    \includegraphics[width=0.85\linewidth]{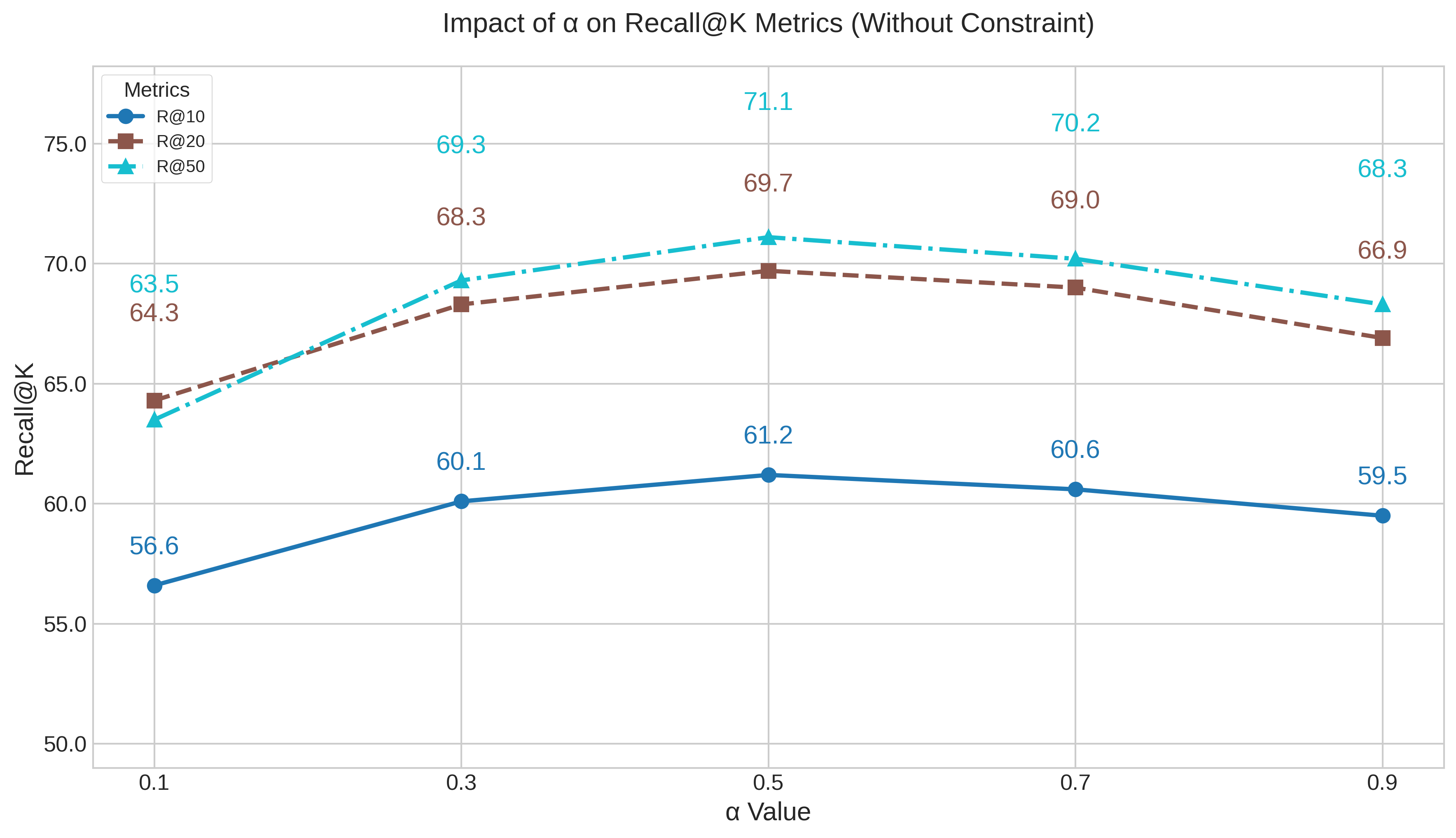}
    \caption{Impact of weight factor $\alpha$ (Without Constraint)}
    \label{fig:alpha_wo}
\end{figure}

\begin{table}
    \centering
    \caption{Impact of LLM on SGCLS}
    \label{tab:ablation_llm}
    \resizebox{0.7\linewidth}{!}{
    \begin{tabular}{ccccccc}
    \midrule
    \multirow{2}{*}{Method} & \multicolumn{3}{c}{With Constraint} & \multicolumn{3}{c}{No Constraint} \\ \cmidrule(lr){2-4} \cmidrule(lr){5-7}
                            & R@10       & R@20       & R@50      & R@10      & R@20      & R@50      \\ \midrule
    w/o LLM      &    38.9        &     40.3       &  40.3         &    47.0       &  57.2         &      63.4     \\ 
    w/ T5       &    51.5        &  53.0          &  53.0       & 59.3          &  68.6    &70.6     \\ \midrule
    SceneLLM     & 53.7 & 55.0            &  55.0            &  61.2         & 69.7           &   71.1                  \\ \midrule    
    \end{tabular}}
    \end{table}

\begin{table}
\centering
\caption{Impact of Features Discretization on SGCLS}
\label{Tab:ablation_discrete}

\resizebox{0.7\linewidth}{!}{
\begin{tabular}{ccccccc}
\midrule
\multirow{2}{*}{Method} & \multicolumn{3}{c}{With Constraint} & \multicolumn{3}{c}{No Constraint} \\ \cmidrule(lr){2-4} \cmidrule(lr){5-7}
                        & R@10       & R@20       & R@50      & R@10      & R@20      & R@50      \\ \midrule
w/o discretization      &   40.4         &       42.6     &   42.6        &    49.0       &   59.0        &  62.3         \\ 
SceneLLM     & 53.7 & 55.0            &  55.0            &  61.2         & 69.7           &   71.1                  \\ \midrule
\end{tabular}}
\end{table}

\noindent\textbf{Impact of weight factor $\alpha$.}
We obtained the optimal weight factor $\alpha$ through grid search. As shown in Fig.\ref{fig:alpha_w} and Fig.\ref{fig:alpha_wo}, under the R@K (K=[10,20,50]) metrics, the model achieved the best performance when alpha was set to 0.5, regardless of constrained or unconstrained settings.

\noindent\textbf{Impact of LLM.} 
In our SceneLLM, we introduce LLM into the SGG task to reason about the spatial-temporal semantic relationships of objects by leveraging its implicit knowledge. 
To evaluate the efficacy of this strategy, we test two variants. In the first variant (\textbf{w/o LLM}), we remove the LLM and directly use $S_{LLM}$ as the input of SGG predictor to generate scene graphs.  In the second variant (\textbf{w/ T5}), we replace our LLM (LLaMA) with a smaller LLM, i.e., T5 \cite{raffel2020exploring}.

The results are shown in Table \ref{tab:ablation_llm}. It can be observed that our method (w/ LLaMA) achieves better performance than the two variants, which demonstrates the effectiveness of leveraging a powerful LLM. In addition, the worse performance of \textbf{w/ T5} shows that a less powerful LLM (T5) may hard to handle relationships in complex scenes.  

\noindent\textbf{Impact of Features Discretization.} 
In our framework, we discretize the encoded latent features of objects to generate discrete word tokens. To validate this approach, we test a variant called \textbf{w/o discretization}, where the continuous visual features are directly used to produce the implicit linguistic signal $S_{LLM}$ as input to the LLM. As shown in Table \ref{Tab:ablation_discrete}, the performance of our SceneLLM is significantly better than \textbf{w/o discretization}. This improvement is because the discretization of features in our framework makes the visual features more "like" human sentences composed of discrete word tokens. As a result, the LLM, pre-trained on human sentences, can more easily interpret these "human sentence-like" visual features.

\noindent\textbf{Impact of the Optimal Transport Scheme.}
In SceneLLM, we design an optimal transport-based scheme to generate a temporally consistent language signal. 
To validate the effectiveness of our strategy, we test three variants:
\textbf{1) w/o OT}, where we remove the Optimal Transport (OT) scheme and directly feed the frame-level tokens ${t_\tau}_{\tau=1}^{T}$ (along with the prompt texts) into the LLM;
\textbf{2) Temporal Convolution (TC)}, where we replace the OT scheme with a temporal convolution module to capture temporal correlations between frame-level tokens before feeding them into the LLM;
\textbf{3) Clustering}, where we substitute the OT scheme with a clustering algorithm to aggregate tokens with temporal correlations and then derive a new codebook $\mathbb{C}^+$.
As shown in Table~\ref{Tab:ablation_ot}, our method outperforms these variants, demonstrating the efficacy of the OT scheme in reconstructing an LLM-friendly language signal from frame-level tokens.

\begin{table}
\centering
\caption{Impact of  the Optimal Transport Scheme on SGCLS}
\label{Tab:ablation_ot}

\resizebox{0.7\linewidth}{!}{
\begin{tabular}{ccccccc}
\midrule
\multirow{2}{*}{Method} & \multicolumn{3}{c}{With Constraint} & \multicolumn{3}{c}{No Constraint} \\ \cmidrule(lr){2-4} \cmidrule(lr){5-7}
                        & R@10       & R@20       & R@50      & R@10      & R@20      & R@50      \\ \midrule
w/o OT       &    50.9        &    52.3        &    52.3       &    58.9       &    67.4       &  70.0         \\ 
TC     &    50.8        &    52.0        &    52.0       &    58.7       &    66.9       &  69.1         \\ 
Clustering      &    50.6        &    51.2        &    51.2       &    57.1       &    65.8       &  68.3         \\ 
\hline
SceneLLM     & 53.7 & 55.0            &  55.0            &  61.2         & 69.7           &   71.1                  \\ \midrule
\end{tabular}}
\end{table}

\begin{table}
\centering
\caption{Impact of the LoRA process on SGCLS}
\label{Tab:ablation_lora}
\resizebox{0.7\linewidth}{!}{
\begin{tabular}{ccccccc}
\midrule
\multirow{2}{*}{Method} & \multicolumn{3}{c}{With Constraint} & \multicolumn{3}{c}{No Constraint} \\ \cmidrule(lr){2-4} \cmidrule(lr){5-7}
                        & R@10       & R@20       & R@50      & R@10      & R@20      & R@50      \\ \midrule
w/o LoRA      &   47.3         &     48.4       &  48.4         &   55.0        &   64.9        &    67.3       \\ 
SceneLLM     & 53.7 & 55.0            &  55.0            &  61.2         & 69.7           &   71.1                  \\ \midrule
\end{tabular}}
\end{table}

\noindent\textbf{Impact of the LoRA Process.} 
In our framework, to enable the LLM to comprehend "action sentences" while preserving its pre-trained weights and the extensive knowledge it has acquired, we fine-tune the model using a LoRA process. To assess the effectiveness of this approach, we also evaluate an alternative variant (\textbf{w/o LoRA}). In this variant, instead of only applying LoRA, all parameters of the LLM, initialized with its pre-trained weights, are subjected to gradient updates during the tuning process in the second training stage of SceneLLM.
As shown in Table \ref{Tab:ablation_lora}, the performance of our SceneLLM is significantly better than the \textbf{\textbf{w/o LoRA}} variant. This result demonstrates that the LoRA process is effective in fine-tuning the LLM to better understand the "scene sentences" while preserving its pre-learned knowledge.


\section{CONCLUSIONS}

In this work, we introduce SceneLLM, a groundbreaking dynamic Scene Graph Generation (SGG) framework that leverages the advanced reasoning capabilities of Large Language Models (LLMs). SceneLLM excels at extracting fine-grained semantic relationships from videos, achieving state-of-the-art performance on the widely used AG benchmark. By utilizing the vast reasoning potential of LLMs, SceneLLM enhances the ability to understand complex interactions and contextual relationships within video data, setting a new standard in the field of SGG. Looking ahead, we plan to extend our research into SGG within open-world environments, as well as explore its applications in 3D scenes. The real world is inherently open and constantly evolving, presenting challenges that closed-world datasets cannot address. Models trained on such limited datasets often face difficulties when encountering novel or unseen situations in practice. This limitation presents significant challenges, particularly for the safe deployment of models in critical applications such as mobile robots and autonomous systems. Therefore, future work will aim to improve model robustness, generalization, and adaptability in the face of unpredictable real-world scenarios, ensuring that these systems can operate safely and effectively in dynamic environments.

\section*{Acknowledgements}
This research did not receive any specific grant from funding agencies in the public, commercial, or not-for-profit sectors.




\bibliographystyle{elsarticle-num}
\bibliography{ref}

\end{document}